\documentclass[11pt]{article}

\usepackage[preprint]{acl}

\usepackage{times}
\usepackage{latexsym}
\usepackage{booktabs}
\usepackage{amsmath}
\usepackage{amssymb}
\usepackage{nameref}

\usepackage[T1]{fontenc}
\usepackage[utf8]{inputenc}

\usepackage{microtype}

\usepackage{inconsolata}

\usepackage{graphicx}
\usepackage{multirow}

\usepackage{cleveref}
\usepackage{soul}
\usepackage{dsfont}
\usepackage{xspace}
\usepackage[dvipsnames]{xcolor}
\usepackage{amssymb}

\newcommand{\defn}[1]{\emph{#1}}

\newcommand{\dirref}[0]{DRT\xspace}
\newcommand{\indirref}[0]{IRT\xspace}
\newcommand{\prosdriv}[0]{PDS\xspace}

\newcommand{\prosodycolor}[0]{cyan}
\newcommand{\wordcolor}[0]{magenta}
\newcommand{\syntaxcolor}[0]{olive}
\newcommand{\mcolor}[0]{gray}

\newcommand{\prosodyunit}[0]{\ensuremath{\textcolor{\prosodycolor}{\mathrm{p}}}\xspace}
\newcommand{\prosody}[0]{\ensuremath{\textcolor{\prosodycolor}{\mathrm{\mathbf{p}}}}\xspace}
\newcommand{\Prosody}[0]{\ensuremath{\textcolor{\prosodycolor}{\mathrm{\mathbf{P}}}}\xspace}

\newcommand{\syntaxunit}[0]{\ensuremath{\textcolor{\syntaxcolor}{\mathrm{s}}}\xspace}
\newcommand{\syntax}[0]{\ensuremath{\textcolor{\syntaxcolor}{\mathrm{\mathbf{s}}}}\xspace}
\newcommand{\Syntax}[0]{\ensuremath{\textcolor{\syntaxcolor}{\mathrm{\mathbf{S}}}}\xspace}

\newcommand{\word}[0]{\ensuremath{\textcolor{\wordcolor}{\mathrm{w}}}\xspace}
\newcommand{\words}[0]{\ensuremath{\textcolor{\wordcolor}{\mathrm{\mathbf{w}}}}\xspace}

\newcommand{\Word}[0]{\ensuremath{\textcolor{\wordcolor}{\mathrm{W}}}\xspace}
\newcommand{\Words}[0]{\ensuremath{\textcolor{\wordcolor}{\mathrm{\mathbf{W}}}}\xspace}

\newcommand{\pos}[0]{\ensuremath{\textcolor{\mcolor}{\mathcal{P}}}\xspace}
\newcommand{\brackets}[0]{\ensuremath{\textcolor{\mcolor}{\mathcal{B}}}\xspace}

\newcommand{\mi}[0]{\ensuremath{\textcolor{\mcolor}{\mathrm{I}}}\xspace}
\newcommand{\ent}[0]{\ensuremath{\textcolor{\mcolor}{\mathrm{H}}}\xspace}
\newcommand{\surp}[0]{\ensuremath{\textcolor{\mcolor}{\iota}}\xspace}
\newcommand{\uncert}[0]{\ensuremath{\textcolor{\mcolor}{\mathrm{U}}}\xspace}

\newcommand{\magn}[0]{\ensuremath{\textcolor{\mcolor}{\mathit{mag}}}\xspace}
\newcommand{\posn}[0]{\ensuremath{\textcolor{\mcolor}{\mathit{pos}}}\xspace}

\newcommand{\Prosodypause}{\Prosody_{\mathtt{pause}}}
\newcommand{\Prosodyduration}{\Prosody_{\mathtt{dur}}}

\title{Using Prosody to Predict Syntactic Structure}
\author{Junghyun Min$^{1}$, \quad Alex Warstadt$^{2}$, \quad Tamar I. Regev$^{3}$, \\ 
    \textbf{ Tiago Pimentel$^{4}$, \quad Ethan Gotlieb Wilcox$^{1}$}\\
  $^{1}$Georgetown University, $^{2}$UC San Diego, $^{3}$MIT, $^{4}$ETH Zürich\\
  {
    \textbf{Correspondence:} \href{mailto:jm3743@georgetown.edu}{\texttt{jm3743@georgetown.edu}}
}
}

\begin{document}
\maketitle
\begin{abstract}
While it is well-established that prosody carries crucial cues for syntactic structure, the degree and nature of correspondence between these two domains remains contested.
We investigate the syntax-prosody interface through an information-theoretic lens, quantifying the interaction between prosodic features and syntactic representations as their \emph{mutual information}.
We provide a general-purpose framework for estimating this quantity over large speech-text corpora using multimodal language models.
Our framework is structure-agnostic and modular, insofar as it can be used to measure the contributions of individual prosodic features or components of structure.
We evaluate the syntax-prosody relationship for two features (word duration and inter-word pauses) across two domains---read audiobooks and spontaneous conversations---both in English.
Our results demonstrate that prosody contains measurable syntactic information, with prosodic features reducing syntactic uncertainty in spontaneous conversations by up to 10.2\%.
Our findings offer new empirical support for several theoretical accounts of the syntax-prosody interface.\footnote{We release code for our experiments and analyses at \url{https://github.com/aatlantise/prosody-syntax-interface}. }
\end{abstract}

\section{Introduction}
\label{sec:intro}
Prosody, or the ``melody'' of speech, carries information that is vital for successful communication.
Individual components of the prosody channel---including word duration, pauses between words, pitch, and intensity---have been shown to encode information about word identity \citep{wolf-etal-2023-quantifying, wilcox-etal-2025-using}, discourse functions \citep{wilson2006relevance, hamlaoui2019acoustic}, and meta-linguistic information \citep{mozziconacci2002prosody, yadavalli2025prosodytextconveycharacterizing}.
Prosody has also long been known to carry vital cues about syntactic structure \citep{Elfner2018syntax}.
Children use prosody to learn the latent structure of their language \citep{soderstrom2003prosodic, christophe2008bootstrapping};
speakers use prosodic cues like pre-boundary lengthening to signal syntactic constituents \citep{KLATT1975vowel, ferreira1993creation, SNEDEKER2003usingprosody}; and listeners rely on prosody to disambiguate between parses during real-time communication \citep{price1991use, Millotte2007phrasal}.

While linguists studying the \textit{syntax--prosody interface} agree that prosody and syntax interact, the manner and quantity of syntactic information carried by prosody remain debated.
Different theories make different predictions about how tightly and in what ways the two are linked.
For example, the Direct Reference Theory (\dirref) posits that prosodic constituency is a direct representation of syntactic constituency, while the Indirect Reference Theory (\indirref) allows mismatches between the two to satisfy phonological well-formedness constraints
(see \Cref{sec:related-work} for more details; \citealp{chomsky1968sound, nespor2007prosodic, richards2016contiguity}).
These theories can be difficult to test due to limitations with current empirical methods.
Much of the empirical evidence relies on case studies that examine individual phenomena in controlled settings \citep{KLATT1975vowel, seidl2013minimal, zubizarreta1998prosody}.
And studies that investigate the syntax--prosody relationship in larger datasets often rely on previous-generation statistical methods \citep{berkeley2011prosody}, or make overly strong simplifying assumptions \citep[e.g. omitting the effects of text context when predicting syntax;][]{anu2014sentence, degano2024speech}.

In light of these limitations, this work proposes a new framework for measuring the relationship between syntax and prosody in large-scale natural corpora.
Building on a line of recent work \citep{wolf-etal-2023-quantifying, regev-etal-2025-time, wilcox-etal-2025-using, yadavalli2025prosodytextconveycharacterizing}, we frame our method using the toolkit of information theory.
We define \defn{syntactic information content} of a prosodic feature as the mutual information (\mi; \citealp{shannon1948mathematical}) between that prosodic feature and a structural representation.
In other words, syntactic information content is the extent to which knowing prosodic features reduces uncertainty over structural interpretations (i.e., syntactic parses) of a sentence.
Our framework is structure-agnostic, insofar as it can be measured with respect to any type of structural representation.
It is also contextual, with predictions conditionable on both the surrounding prosody or text context.
And it is modular, insofar as it can be used to measure the contributions of individual prosodic features or components of a structural representation, such as phrase-boundary locations or POS tags.

We implement a computational pipeline to measure the syntactic information content in two English text--speech corpora.
We use modified T5 encoder-decoder transformer models \citep{raffel2020t5} to predict linearized syntactic parses, conditioned on either just text, just prosody, or prosody and text (see \Cref{sec:model-arch}).
We run a series of experiments to investigate: 
(1) how much syntactic information content is carried by individual prosodic features (word duration and inter-word pauses);
(2) how syntactic information content changes across speech genre (planned vs.\ spontaneous speech);
(3) how much information prosody carries beyond what is carried by the segmental information (i.e., words); and
(4) what parts of the syntactic structure prosody contains information about (phrasal categories vs.\ phrase boundaries).

Our results indicate that prosody carries information about syntactic structure, with duration and pause reducing uncertainty over syntactic representations by up to 10.2\%.
These features primarily signal boundary location, as opposed to phrase category identity, and carry more information in spontaneous over planned speech, as a proportion of the total structural uncertainty in each genre.
We do not find evidence that prosody contains additional information about structure beyond what is carried by words; the syntactic information in duration and pause is redundant with that contained in text.
% \alex{Doesn't this contradict the previous 2 sentences?}
We interpret our results as being consistent with the Prosodic Bootstrapping theory of syntactic acquisition \citep{soderstrom2003prosodic, christophe2008bootstrapping} and the Indirect Reference approach to the syntax--prosody interface \citep{selkirk2011syntax, seidl2013minimal}.
More broadly, we argue that our methods can go beyond previous-generation statistical approaches to reveal and quantify new statistical dependencies at the prosody-syntax interface.
%Despite some limitations (see \Cref{sec:limitations}), our studies use large-scale corpora and LM-based estimation methods to reveal and quantify statistical dependencies that are otherwise difficult to uncover.
% \alex{Leveraging information theory and modern language models is not a selling point on its own. The main selling point is that these methods reveal and quantify obscure statistical relationships that other approaches cannot hope to find.}
% most technically sophisticated\tiago{why is it the most technically sophisticated method?} large-scale corpus evidence\tiago{a bit of a mouthful "most technically sophisticated large-scale corpus evidence". this sentence could probably be rewritten more clearly imo} in the literature to date about how prosody and syntax interact.

\section{The Syntax--Prosody Interface}
\label{sec:related-work}

\subsection{Theoretical Linguistics}

Accounts of the syntax--prosody interface in theoretical linguistics often focus on the interplay between prosodic and syntactic well-formedness constraints.
Different theories postulate different degrees of causal linking, which we argue makes predictions about the informational relationship between prosody and syntactic structure.

\paragraph{Direct Reference Theory (\dirref).}
\dirref proposes that prosodic rules don't employ separate phonological structures, but are formulated with respect to the underlying syntactic parses \citep{chomsky1968sound, cooper1980syntax, kaisse1985theory}.
This is supported by phonetic evidence showing that speakers consistently produce pre-boundary lengthening at the edges of syntactic constituents \citep{KLATT1975vowel}. 
Under this view, prosody is isomorphic to syntax; mismatches between syntax and prosody are attributed to variance in underlying syntactic operations, such as extraposition \citep{wagner2015phonological}.
We interpret \dirref to predict a very high degree of syntactic information content, with a strict interpretation suggesting that prosodic information should come close to reducing all uncertainty about a sentence's syntactic parse.

\paragraph{Indirect Reference Theory (\indirref).}
\indirref argues that syntax and phonology are autonomous domains that communicate via a mediating mechanism \citep{nespor2007prosodic, seidl2013minimal}.
Match Theory \citep{selkirk2011syntax} formalizes this as a set of constraints that strive for isomorphism between syntactic and prosodic constituents, though this mapping can be overridden when certain phonological well-formedness constraints are ranked to outweigh isomorphism by Optimality Theory\footnote{Optimality Theory considers grammatical rules as ranked and hierarchical; it selects the linguistic form that violates the least high-ranked rules.} \citep{prince2004optimality}.
\indirref predicts a lower amount of information overlap; this is due to the fact that prosody and syntax each operate within its own framework of well-formedness constraints, with one causally influencing the other only when the two are in conflict.

\paragraph{Prosody-Driven Syntax (\prosdriv).}
\prosdriv argues that the interface is bidirectional: syntactic well-formedness can drive prosodic choices, but prosodic well-formedness can also influence the syntax of a sentence.
Certain syntactic operations like wh-movement, heavy NP shift, or argument scrambling are triggered specifically to satisfy prosodic requirements such as nuclear stress \citep{zubizarreta1998prosody}, or contiguity \citep{richards2016contiguity}.\footnote{\emph{Nuclear stress} \citep{zubizarreta1998prosody} describes how constituents move to align the discourse focus with the sentence's primary phonological peak 
%(e.g., shifting a subject to the end of a clause to mark it as ``new'' information)
; \emph{contiguity} \citep{richards2016contiguity} posits that movement occurs to satisfy phonological phrasing requirements between a head and its dependent.}
%A classic case is Heavy NP Shift, where an internally complex object is displaced to the right periphery to prevent a phonological ``break'' between the verb and its subsequent modifiers.
We take \prosdriv to make no strong predictions about the total degree of informational overlap, but rather about differences between genres, specifically spontaneous vs. planned speech.
Because in planned (read-out-loud) speech, the syntax of a sentence is fixed, it cannot be modulated by prosodic choices.
However, in spontaneous speech, as prosodic choices can cause changes in syntax, prosody should provide more information about structure.
We therefore interpret \prosdriv to predict a higher degree of syntactic information content in spontaneous, as opposed to planned speech.

\paragraph{Prosodic Bootstrapping.}
%Before full acquisition, infants also rely on prosodic cues to infer word and phrase boundaries \citep[prosodic bootstrapping; ][]{soderstrom2003prosodic, christophe2008bootstrapping}.
Prosodic Bootstrapping is a theory of language acquisition that posits that infants use prosodic information to make inferences about the syntactic structure of a sentence, even when they are unaware of its semantics, or when it includes words that they haven't yet learned.
Proponents of the theory find evidence that infants as young as 6 months old are sensitive to prosodic markers of phrasal units \citep{soderstrom2003prosodic} and that they use this prosodic constituency information to constrain lexical segmentation \citep{christophe2008bootstrapping}.
Theories of prosodic bootstrapping predict non-zero syntactic information content in prosody, which is the basis of the bootstrapping.

% \paragraph{Speech Planning.}
% Related work in psycholinguistics also consider other cognitive constraints that may affect prosody realization, suggesting a more nuanced mechanism in the syntax-prosody interface \citep{chafe1994discourse, croft1995intonation}.
% One such factor is speech planning.
% \citet{bannon2026syntactic} investigate the alignment between prosodic and syntatic phrase boundaries, but find that rate of speech and speech planning is also an important factor affecting prosodic boundary insertion.
% According to their findings, the more the speech is planned, the higher the alignment between prosodic and syntactic phrase boundaries.
% However, it is unclear whether reading constructed sentences out loud behaves similarly with online planning during speech production.

\subsection{Computational \& Info-Theoretic Models}

Prosodic information has long been used as featural inputs to structure predictions in NLP systems \citep{kahn-etal-2005-effective, anu2014sentence, berkeley2011prosody, cho2022leveraging, min-etal-2025-punctuation}, and vice versa \citep{koehn2000improving, dhamne-etal-2025-predicting}.
This study builds on a line of work quantifying the informational redundancy between prosody and non-prosodic linguistic channels.
This work framed informational redundancy as mutual information \citep{pimentel-etal-2020-information} and used large language models to measure this quantity between prosody and segmental information operationalized as text \citep{wolf-etal-2023-quantifying}, pitch, and word identity \citep{wilcox-etal-2025-using}, as well as to assess the temporal dynamics of the overlap \citep{regev-etal-2025-time}.
One shortcoming of this previous work was that because the computational pipelines made predictions over continuous-valued prosodic features, estimates (of entropy) were not bounded from below, limiting the interpretation of results.
One recent contribution \citep{yadavalli2025prosodytextconveycharacterizing} proposed a solution to this problem by using prosody to measure auxiliary discrete-valued tasks, such as emotion classification or sarcasm detection.
The present work builds on this contribution, using syntax as our auxiliary task.

\section{Modeling Framework}
\label{sec:entropy-mi}

In this section, we describe our formal framework for estimating the syntactic information carried by prosodic features, which we frame as the mutual information between prosody and syntax.
To do so, we treat syntax (\Syntax), prosody (\Prosody), and text (\Word) as random variables (RVs).
A number of operationalizations and design choices follow.

\subsection{Our Random Variables}
\label{sec:random-variables}

% \paragraph{Sentences as a sequence of symbols.}

%We now introduce the random variables we discuss in this work.
First, we consider a word, \word, to be an element from a 
% (potentially infinite)
vocabulary $\Sigma$.
% \tiago{Does this need to be finite? Words are arguably infinite. I changed this. Please undo, if this breaks anything else.}
A \defn{text} is then a sequence of words, $\words \in \Sigma^*$, which we can write as: $\words = [\word_1, \word_2, \cdots, \word_N]$.
% We use \Word and \Words to denote, respectively, word- and text-valued random variables.
% \Words is a discrete-valued random variable that ranges over these sequences.
% 
% \paragraph{Syntactic structure as linearized constituency parse.}
% 
Second, we consider a syntactic unit, $\syntaxunit$, as an element from the set: $\Upsilon = \pos \cup \brackets$, where \pos is a finite set of symbols representing phrasal category tags (e.g., \texttt{NN}, \texttt{SBAR}), and \brackets is a set containing an opening and closing bracket.\footnote{In some of the experimental settings, we ignore phrasal categories, and thus $\Upsilon = \brackets$.}
We then represent the \defn{syntax}, $\syntax \in \Upsilon^*$, of a text as a sequence of symbols drawn from $\Upsilon$, e.g.,: \texttt{(ROOT (S (NP PRP) (VP VBZ (NP DT NN NN))))}.
Finally, we treat a prosodic unit, $\prosodyunit \in \mathbb{R}$, as a real-valued feature.
In our experiments, this corresponds to word-level duration and pause length (measured in milliseconds; $ms$), but this could correspond to any prosodic feature.
We then represent the \defn{prosody}, $\prosody \in \mathbb{R}^{d}$, of a text as a sequence of prosodic units.

We assume that there exists an unknown ground-truth distribution over the three variables above, $p(\words, \syntax, \prosody)$, and that for any tuple $\langle\words, \syntax, \prosody\rangle$ with non-zero probability, $\mathrm{length}(\words)=\mathrm{length}(\prosody) = d$.
% \alex{Two questions: What is $d$ (is this a definition of $d$)? And why isn't this probability 1 if these are word-level prosodic features?}
% \hyun{with non-zero probability modifies tuple! $d$ is the length of the input vector \Prosody. Let me rewrite this.}
We use \Words, \Syntax, and \Prosody to denote, respectively, sentence-level text-, syntax- and prosody-valued random variables.\looseness=-1

\subsection{Measuring Information Content}
\label{sec:info-content}

%As described above, we wish to measure the information content that prosody carries about syntax.
We operationalize the information that prosody carries about syntax as the mutual information: $\mi(\Prosody, \Syntax)$, following prior work \citep{wolf-etal-2023-quantifying,wilcox-etal-2025-using,regev-etal-2025-time,yadavalli2025prosodytextconveycharacterizing}.
% between random variables that represent prosody and syntax, \Prosody and \Syntax.
% Estimating this mutual information is non-trivial;
To estimate mutual information, 
we follow \citet{pimentel-etal-2019-meaning}'s two-step process of estimation:
We first decompose \mi into the difference between an unconditional and a conditional entropy.
We then use probabilistic models to upperbound each of these entropies (as further described in \Cref{sec:entropy-estimates}).
With the ``good mixed-pair assumption,'' \citep{nair2007entropymixturesdiscretecontinuous,beknazaryan2019mutual}, the mutual information decomposes as:\footnote{This assumption requires that $\ent(\Prosody \mid \syntax)$ be well-defined and finite for all values of $\syntax$.}
\begin{align}
    \mi( \Prosody, \Syntax) = \ent(\Syntax) - \ent(\Syntax \mid \Prosody).
\end{align}
% % 
% In \citeauthor{pimentel-etal-2020-information}'s framework, each entropy is estimated separately.
% We describe our entropy estimation in \Cref{sec:entropy-estimates}.

% \paragraph{Disentangling effects beyond word identity.}

% Beyond estimating this mutual information, however,
In written or transcribed text, however, words carry substantial information about syntactic structure.
It is thus a natural question to ask whether prosody carries information \emph{above and beyond} what is carried by word identities.
Beyond estimating the mutual information above, 
% To disentangle the separate effects of prosody and text, 
we thus additionally estimate the mutual information between prosody and syntax conditioned on word identities,  $\mi(\Prosody, \Syntax \mid \Words)$.
We decompose it using the same logic as above:
\begin{align}
    \mi(\Prosody, \Syntax \mid \Words) = \ent(\Syntax \mid \Words) - \ent(\Syntax \mid \Words, \Prosody).
\end{align}
If this value is positive, it means that prosody carries additional information about syntactic structure not carried by the text.

\subsection{Entropy Estimation}
\label{sec:entropy-estimates}

% MACROING FOR THIS SECTION
\newcommand{\params}[0]{\ensuremath{\textcolor{\mcolor}{\mathbf{\theta}}}\xspace}
\newcommand{\data}[0]{\ensuremath{\textcolor{\mcolor}{\mathcal{D}}}\xspace}

%Similarly to the mutual information, the entropy is a non-trivial quantity to estimate.
% However, given any probabilistic model $p_{\params}()$, we can upperbound it 
We follow \citet{pimentel-etal-2019-meaning} in estimating entropy via a cross-entropy upper-bound.
Let $\ent_{\params}(\Syntax)$ be the \defn{cross-entropy} between any distribution $p_{\params}$ and the ground truth distribution $p$.
We can use it to upperbound the entropy via the inequality: 
$\ent_{\params}(\Syntax) \geq \ent(\Syntax)$.
The cross-entropy is 
% , where the learned distribution is parameterized by a neural network.
defined as the expected surprisal of $p_{\params}$ taken with respect to $p$, where \defn{surprisal} under $p_{\params}$ is defined as $\surp(x) = - \log p_{\params}(x)$.
As $p$ is unknown, we take a Monte Carlo estimate of the cross-entropy.
Given a model $p_{\params}$ and a dataset $\{\words^{n}, \syntax^{n}, \prosody^{n}\}_{n=1}^{N}$ sampled according to $p$, we can estimate it as:
% 
% follow \citet{wolf-etal-2023-quantifying} and empirically estimate it using resubstitutive sampling \citep{ahmad1976nonparametric, hall1993estimation}:
% Given our learned distribution $p_{\params}$ and a held-out dataset with $N$ datapoints, the cross entropy is the average surprisal of the test items:
% 
\begin{align}
        \ent_{\params}(\Syntax) 
        &= - \sum_{\syntax \in \Upsilon^*} p(\syntax) \log p_{\params}(\syntax) \\
        &\approx - \frac{1}{N} \sum_{n = 1}^N \log p_{\params}(\syntax^{n}), \texttt{where } \syntax^{n} \sim p(\syntax). \nonumber
\end{align}
% 
% \noindent where $\syntax^n$ is the $n^{th}$ test item. 
Analogous equations exist for the conditional entropies, e.g., $\ent_{\params}(\Syntax \mid \Prosody)$, but using the conditional distribution instead, e.g., $p_{\params}(\syntax \mid \prosody)$. 
Notably, the difference between the cross-entropy and entropy, e.g.,  $\ent_{\params}(\Syntax) - \ent(\Syntax)$, is given by the Kullback-Leibler divergence \citep{kullback1051information} between $p_{\params}$ and $p$. 
This implies that, the more similar $p_{\params}$ and $p$ are, the better our estimators will be.
%We now describe how we model $p_{\params}$.

\paragraph{Estimating $p_{\params}(\syntax)$.}
\label{exp1}
We train a language model from scratch on a corpus of linearized constituency parses until convergence.
No conditioning information (e.g., text or prosody) is provided, and this model is thus composed of a simple decoder-only autoregressive component.
% The resulting cross-entropy loss is a token-level metric; we multiply the average token-level entropy by the average length of parses in the test set to obtain the average sequence-level entropy.\tiago{I think this is an implementational detail which could be deleted imo. Maybe just say instead somewhere that "All our reported cross-entropies, e.g., $\ent_{\params}(\Syntax)$, are given per sentence, without any type of token-level normalisation."}

\paragraph{Estimating conditional distributions $p_{\params}(\syntax \mid \words)$, $p_{\params}(\syntax \mid \prosody)$, and $p_{\params}(\syntax \mid \words, \prosody)$.}
We fine-tune an encoder-decoder language model to predict syntactic sequences (in the decoder) given a text sequence, a prosody sequence, or both sequences as input (in the encoder).
While we train the decoder component of this model from scratch, the text encoder and cross-attention layers are loaded with pre-trained weights;\footnote{From \texttt{t5-base}; see details in \Cref{sec:model-arch}.\looseness=-1} this model can thus make use of text information it has learned during pre-training.
All the cross-entropies reported here are over sentence-level variables, requiring no token-level normalization.

\section{Methods}
\label{sec:methods}
%Section \ref{sec:entropy-mi} describes how we measure syntactic information content via entropy estimation.
The basis of our entropy estimation is multimodel encoder-decoder LMs which we train to predict syntactic structure from text input, prosody input, or no input.
In this section, we detail our implementation pipeline, focusing on how we train LMs to predict (linearized) syntactic structures.
We discuss model architectures in Section \ref{sec:model-arch} and data (including datasets, syntactic-structure operationalisation, and statistical testing) in Section \ref{sec:datasets}.\looseness=-1

\subsection{Encoder-Decoder Model}
\label{sec:model-arch}

\newcommand{\ME}[0]{\ensuremath{\mathrm{ME}}\xspace}
\newcommand{\PE}[0]{\ensuremath{\mathrm{PE}}\xspace}

\begin{figure*}
    \centering
    \includegraphics[width=\linewidth]{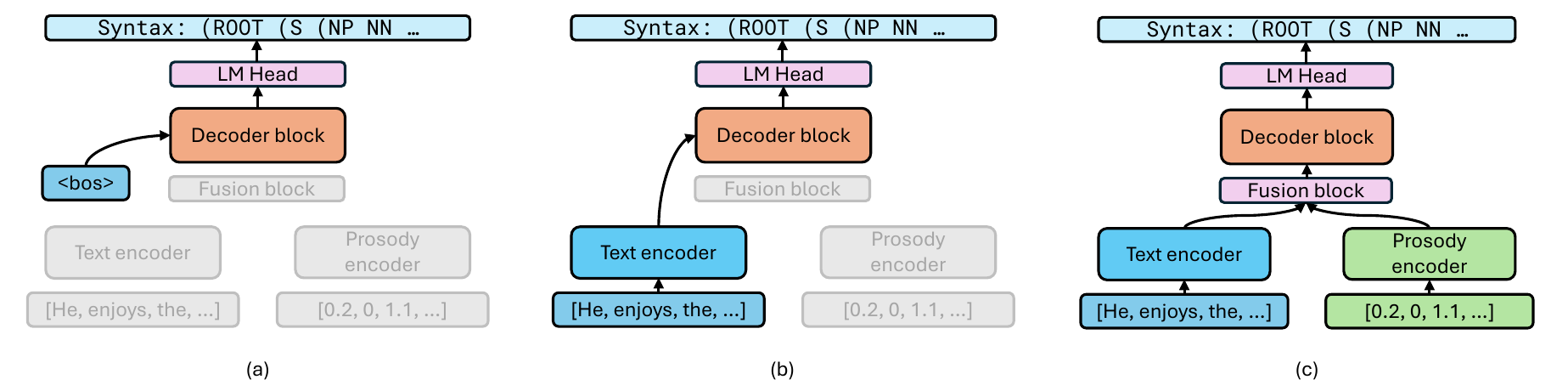}
    \caption{Architecture of our encoder-decoder model. (a), (b), and (c) are used for estimations of $\ent(\Syntax)$, $\ent(\Syntax \mid \Words)$, $\ent(\Syntax \mid \Words, \Prosody)$, respectively.
    To estimate $\ent(\Syntax \mid \Prosody)$, we use an architecture symmetric to (b), but with an active Prosody encoder instead.}
    \label{fig:model-arch}
\end{figure*}

Our sequence-to-sequence model is based on the base-sized encoder-decoder T5 architecture \citep{raffel2020t5}, with several additions, illustrated in Figure \ref{fig:model-arch}.
In order to process prosodic inputs, we construct a new \defn{prosody encoder} from scratch, which differs from T5's text encoder in several respects:
It is shallower and narrower, with 2 layers, 2 heads per layer, and a hidden dimensionality of $8$.
Rather than learning an embedding matrix during training, we used a fixed embedding schema, with two sinusoidal functions used to construct an encoding of the prosodic feature's magnitude \ME and absolute position encoding \PE, given below.
The two encodings are then concatenated before being passed to the transformer stack:
\begin{align}
\small
    \ME_{(\magn, d)}=
    \begin{cases}
        \sin\left(\frac{\magn}{20000^{2i/D}}\right) & \text{if } d = 2i \\
        \cos\left(\frac{\magn}{20000^{2i/D}}\right) & \text{if } d = 2i + 1 \\
    \end{cases}
\end{align}
\begin{align}
\small
    \PE_{(\posn, d)}=
    \begin{cases}
        \sin\left(\frac{\posn}{10000^{2i/D}}\right) & \text{if } d = 2i \\
        \cos\left(\frac{\posn}{10000^{2i/D}}\right) & \text{if } d = 2i + 1 \\
    \end{cases}
\end{align}
Where \magn is the prosodic feature binned into increments of $10ms$. The final hidden state of the encoder is projected to match the hidden state dimensions of the text encoder to allow for stable fusion via cross-attention to the decoder stack.

For estimates of $p_{\params}(\syntax \mid \prosody, \words)$ that condition on both prosody and text, we add a cross-attention layer to fuse inputs across modalities.
The cross-attention is performed with the layer's query as text representation and key and value as prosody representations.
Dropout and layer normalization are applied to the resulting fused representations.
We add part-of-speech labels as additional tokens to the tokenizer, with one (and only one) token per POS label; we assign each of these new tokens a randomly initialized embedding.
These additional tokens are used strictly for decoding (i.e., for predicting syntax) and are not used to tokenize the input text $\words$.
% We randomly initialize the embeddings of these new POS tokens.
The text encoder, embedding layers, and cross-attention layers are loaded with available \texttt{t5-base} pre-trained weights \citep{raffel2020t5}.
The decoder, prosody encoder, fusion layers, and LM head are randomly initialized and trained from scratch to predict syntactic parse tokens.
We report hyperparameters in Appendix \ref{sec:hyperparams}.

\subsection{Datasets}
\label{sec:datasets}

Previous work using methodologies similar to ours has focused on prepared, read-out-loud speech \citep{wolf-etal-2023-quantifying, regev-etal-2025-time, wilcox-etal-2025-using}.
To investigate how prosody may vary across speech styles, we consider two datasets in our experiments:
\defn{LibriTTS} is a dataset of audiobooks and their gold transcriptions,\footnote{This may underestimate prosody's information content; see \nameref{sec:limitations}.} derived from the LibriSpeech corpus \citep{panayotov2015librispeech}, with audiobooks from LibriVox,\footnote{\url{https://librivox.org/}} and the original material from Project Gutenberg.\footnote{\url{https://www.gutenberg.org/}}
We use the aligned version of its \emph{clean} subset using the Montreal Forced Aligner \citep[MFA; ][]{McAuliffe2017MFA}, which comprises 44k sentences.
Our second dataset is \defn{CANDOR} (Conversation: A Naturalistic Dataset of Online Recordings), which consists of 1656 naturalistic, unscripted conversations between strangers that were collected over Zoom, whose transcription pipeline includes automatic speech recognition (ASR).
We preprocess the dataset via the pipeline from \citet{Clark2025Surprisal} and obtain pause and duration information, again by aligning them with MFA. Out of 750k sentences that are 5 words or longer, we filter for those that are not non-initial fragments due to speaker backchannel interruptions, resulting in 298k sentences.
Punctuation marks are removed from the input text (as well as associated syntactic parse), as they are not part of word identities and carry syntactic information \citep{chafe1988punctuation}.
We report the dataset statistics in \Cref{tab:dataset_stats}.

\begin{table}[t]
\centering
\small
\begin{tabular}{lcc}
\toprule
\textbf{Metric} & \textbf{CANDOR} & \textbf{LibriTTS} \\ \midrule
Dataset size (sentences) & 298k & 44k \\
Avg. words per sentence & 13.2 & 17.5 \\
Avg. tokens per full parse & 64 & 76 \\ 
Avg. tokens per brackets & 33 & 38 \\
\bottomrule
\end{tabular}
\caption{Dataset statistics for CANDOR (spontaneous speech) and LibriTTS (planned speech).}
\label{tab:dataset_stats}
\end{table}

\paragraph{Syntactic parses.}
\label{sec:parses}

Although our method could be employed with any structural representation, for our experiments, we represent syntactic structure as linearized constituency parses, following the Penn Treebank format.
As we are interested in what types of syntactic information prosody contains, we generate two types of constituency trees, corresponding to differing levels of parse granularity:
For our \defn{full parse} condition, we obtain silver parse labels on our two datasets using Stanza \citep{qi-etal-2020-stanza}.\footnote{Stanza employs the Penn Treebank framework. The framework processes disfluencies in its spoken subsets like Switchboard \citep{godfrey1992switchboard} by isolating them and annotating the remaining sentence. Stanza's training data was not sourced from the spoken genres.}
From the silver parses, we remove any punctuation marks.
For our \defn{bracket-only} condition, we strip out all phrase category information, leaving only brackets that indicate the hierarchical structure of the sentence. 
Because we use autoregressive language models to estimate $p_{\params}(\syntax)$, it's possible that the model can assign probability to non-valid syntactic parses.
Thus, a certain amount of our probability can \emph{leak}.
To address this concern, we estimate the amount of leaked probability in Appendix \ref{sec:leakage}.

\paragraph{Prosodic features.}
We measure the syntactic information content of word duration and inter-word pauses.
We choose these features because they have been argued to provide cues for syntactic structure \citep{KLATT1975vowel, RATNER1986303}.
Duration is the difference between the word's onset and offset time, as taken from MFA-aligned word boundaries.
Inter-word pauses are taken as the difference between a word's offset and the onset of the next word; again, taken from MFA-aligned boundaries.
Duration is normalized by the number of syllables in a word, using syllable data from CELEX \citep{baayen1995celex2}.
Both measures are taken in milliseconds, but chunked into $10ms$ bins in our prosody encoder (e.g. values $14ms$ and $18ms$ are represented in the same $10-19ms$ bin).
% \tiago{Does this mean that prosody and words can have different lengths?}

\begin{figure*}
    \small
    \centering
    \includegraphics[width=\linewidth]{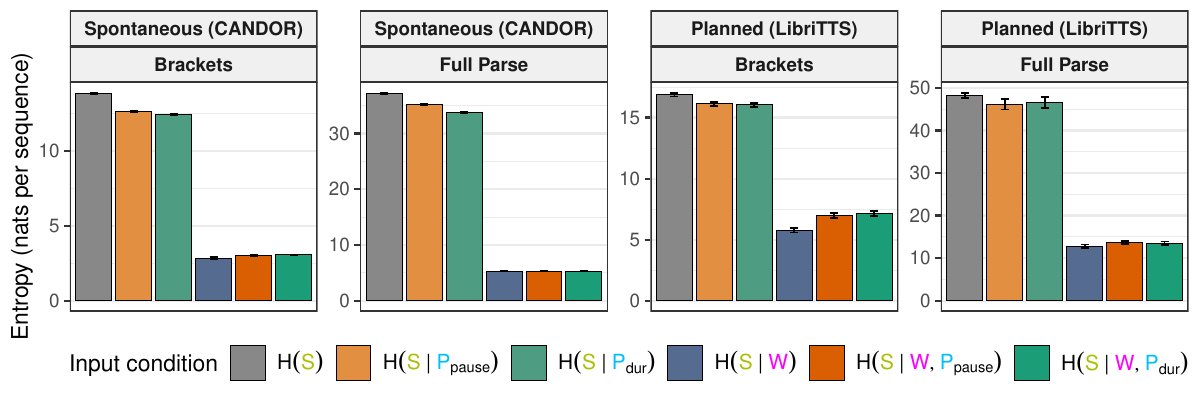}
    \caption{Estimated entropy in predicted syntactic structure given context across conditions.}
    \label{fig:bar_overall_results}
\end{figure*}

\begin{table*}[htb]
\centering
\small
\begin{tabular}{llcccccc}
\toprule
\textbf{Genre} & \textbf{Parse Type} & $\ent(\Syntax)$ & $\ent(\Syntax\mid\Words)$ & $\ent(\Syntax\mid \Prosodypause)$ & $\ent(\Syntax\mid \Words, \Prosodypause)$ & $\ent(\Syntax\mid \Prosodyduration)$ & $\ent(\Syntax\mid \Words,\Prosodyduration)$ \\ \midrule
\multirow{2}{*}{Spontaneous} & Brackets & 13.9 & 2.83 & 12.7 & 3.03 & 12.4 & 3.06 \\
 & Full parse & 37.2 & 5.25 & 35.1 & 5.27 & 33.7 & 5.30 \\ \midrule
\multirow{2}{*}{Planned} & Brackets & 16.9 & 5.76 & 16.1 & 6.97 & 16.0 & 7.13 \\
 & Full parse & 48.2 & 12.8 & 46.2 & 13.6 & 46.6 & 13.5 \\ \bottomrule
\end{tabular}
    \caption{Syntactic entropy (nats per sequence) across datasets and input conditions. LibriTTS is a dataset of planned read-out-loud speech, while CANDOR is a dataset of naturalistic, spontaneous conversations.}
    \label{tab:entropy_folds}
\end{table*}

\begin{table*}[ht]
\centering
\small
\setlength{\tabcolsep}{3pt}
\begin{tabular}{llrrrrrrrrrr}
\toprule
 & & \multicolumn{2}{c}{$(\Syntax, \Prosodypause)$} & \multicolumn{2}{c}{$(\Syntax, \Prosodyduration)$} & \multicolumn{2}{c}{$(\Syntax, \Words)$} & \multicolumn{2}{c}{$(\Syntax, \Prosodypause \mid \Words)$} & \multicolumn{2}{c}{$(\Syntax, \Prosodyduration \mid \Words)$} \\
\cmidrule(lr){3-4} \cmidrule(lr){5-6} \cmidrule(lr){7-8} \cmidrule(lr){9-10} \cmidrule(lr){11-12}
\textbf{Genre} & \textbf{Parse Type} & \multicolumn{1}{c}{\mi} & \multicolumn{1}{c}{\uncert} & \multicolumn{1}{c}{\mi} & \multicolumn{1}{c}{\uncert} & \multicolumn{1}{c}{\mi} & \multicolumn{1}{c}{\uncert} & \multicolumn{1}{c}{\mi} & \multicolumn{1}{c}{\uncert} & \multicolumn{1}{c}{\mi} & \multicolumn{1}{c}{\uncert} \\ \midrule
\multirow{2}{*}{Spontaneous} & Brackets & 1.20 & 8.64\% & 1.42 & 10.2\% & 11.0 & 79.6\% & $-$0.20 & $-$7.14\% & $-$0.24 & $-$8.32\% \\
 & Full parse & 2.05 & 5.52\% & 3.45 & 9.27\% & 31.9 & 85.9\% & $-$0.02 & $-$0.42\% & $-$0.06 & $-$1.07\% \\
\midrule 
\multirow{2}{*}{Planned} & Brackets & 0.75 & 4.44\% & 0.84 & 4.99\% & 11.1 & 65.9\% & $-$1.21 & $-$21.0\% & $-$1.37 & $-$23.8\%\\
 & Full parse & 2.06 & 4.27\% & 1.65 & 3.43\% & 35.4 & 73.5\% & $-$0.89 & $-$6.95\% & $-$0.73 & $-$5.70\% \\
\bottomrule
\end{tabular}
    \caption{Syntactic information content in input feature as mutual information \mi and uncertainty coefficient \uncert. All positive \mi values are statistically significant ($p<0.001$).}
    \label{tab:mutual-info}
\end{table*}

\paragraph{Statistical tests.}
\label{sec:stat-tests}
We test the statistical difference in sentence-level surprisal across conditions using our cross-validation splits.
To this end, we use a Wilcoxon signed-rank test \citep{wilcoxon1945individual}, as sequence-level surprisal values are paired across conditions. 
Notably, when applied to our setting, this test evaluates whether our MI estimates are positive significantly more often than expected by chance, and not if its average value is larger than zero across splits---thus being more resistant to outlier negative estimation errors (note that \mi cannot, by definition, be negative).

\section{Results}
\label{sec:results}

%We report our experimental results and answer our research questions introduced in Section \ref{sec:intro}.
\Cref{tab:entropy_folds} and \Cref{fig:bar_overall_results} outline our entropy estimates $\ent(\text{\Syntax})$, $\ent(\text{\Syntax}\mid\text{\Prosody})$, $\ent(\text{\Syntax}\mid\text{\Words})$, $\ent(\text{\Syntax}\mid\text{\Prosody,\Words})$ across input conditions and domain.
\Cref{tab:mutual-info} shows mutual information \mi and uncertainty coefficient \uncert, defined as the ratio between \mi and $\ent(\Syntax)$.
Prosody is labeled as $\Prosodypause$ and $\Prosodyduration$, when representing, respectively, pause  and duration information.

Overall, pause and duration carry similar amounts of syntactic information content in each condition.
As noted by previous language modeling work \citep[e.g.,][]{godfrey1992switchboard, zen2019librittscorpusderivedlibrispeech}, naturalistic, conversational speech has lower syntactic uncertainty across all conditions: both full and bracket parses, and both with and without text and prosody input.
This may be due to a number of factors, including CANDOR's larger dataset size that would allow a greater degree of convergence (298k vs. 44k samples), or the longer average sentence length in LibriTTS compared to CANDOR (17.5 vs. 13.2 words per sentence) and thus longer average syntax parses (76 vs. 64 tokens).
It is also possible that the syntactic structures found in naturalistic, conversational speech are more predictable than those in recorded audiobooks.\looseness=-1

\paragraph{Answering RQ1: How much information about syntax is carried by word duration and pause?}
Our experiments show that both pause and duration carry measurable syntactic information, in both spontaneous and planned speech.
Our statistical tests show that the sequence-level syntactic information $\mi(\Syntax, \Prosodypause)$ and $\mi(\Syntax, \Prosodyduration)$ is significantly greater than zero in all input conditions ($p < 10^{-4}$).
We find that pause can contain up to 2.06 nats of syntactic information, while duration can contain up to 3.45 nats.
Words alone carry about 30 nats of syntactic information, suggesting that pause and duration carry around 10\% of the information load that words do about syntax.

\paragraph{Answering RQ2: Do prosodic features carry syntactic information beyond what is carried by words?}
We find that after controlling for the syntactic information already carried by words, both pause and duration carry little or no significant measurable syntactic information in planned speech.
We hypothesize this may be due to the sparsity of sentences and constructions where syntax could be disambiguated with prosody.
Our framework may also underestimate the syntactic information beyond what is carried by words, a possibility which we discuss in greater detail in \nameref{sec:limitations}.
For the rest of the results section, we focus on the individual contributions of prosody (that is, $\mi(\Syntax,\Prosody)$, rather than on $\mi(\Syntax,\Prosody\mid \Words)$).

\paragraph{Answering RQ3: How does information content vary across speech styles?}
Duration carries more syntactic information in spontaneous than in planned speech, when quantified both in terms of total mutual information and uncertainty reduction.
Pauses have higher total mutual information in spontaneous speech for our full parse condition, but less in our bracket-parse condition.
However, when quantified in terms of uncertainty reduction, pauses contribute to a greater reduction in spontaneous speech in both conditions.
The overall trend in our results is clear: in the datasets we use, prosody carries more syntactic information in spontaneous, rather than planned, speech.
Sentence length may be a contributing factor in the difference, although utterance length differences could be thought of as an inherent part of the differences between these two genres.
An analysis in \Cref{sec:follow-up-analysis} controlling for the length shows that the cross-genre difference remains significant after controlling for the length distribution.

\paragraph{Answering RQ4: What parts of the syntactic structure does prosody contain information about?}
In both spontaneous and planned speech, phrase boundaries (represented by brackets) account for the majority of syntactic information content in prosody.
Phrase boundaries contain up to around 50\% of the syntactic information content in duration; 1.42 out of 3.45 nats in spontaneous speech and 0.84 out of 1.65 nats in planned speech.
They contain up to 60\% of the syntactic information content in pause; 1.20 out of 2.05 nats in spontaneous speech and 0.75 out of 2.06 nats in planned speech.
This finding is in line with well-established phonological phenomena that connect prosodic features to phrase-structural boundaries, for example, pre-boundary lengthening \citep{KLATT1975vowel, ferreira1993creation}.

\section{Sentence Features and Disfluencies}
\label{sec:follow-up-analysis}

As described in \Cref{sec:entropy-estimates}, our reported cross-entropies are over sentence-level variables, without token-level normalization.
However, at the sentence level, it's possible that estimates of syntactic information content interact with sequence-level properties, such as length and syntactic complexity.
Thus, we perform three follow-up analyses, all using pointwise estimates of syntactic information content---$\surp(\syntaxunit) - \surp(\syntaxunit,\prosodyunit)$, i.e., the syntactic information \emph{of a single sequence}---which are averaged across a dataset to obtain $\mi(\Syntax, \Prosody)$ values in the previous sections.
%considering that pointwise estimates of syntactic information content to be sentence-level analogs of $\mi(\Syntax \mid \Prosody)$---that is, $\surp(\syntaxunit) - \surp(\syntaxunit\mid\prosodyunit)$.
%One analyzes the relationship between sentence-level syntactic information content and sentence features, another analyzes whether sentence features affect trends across genres, and the last analyzes the effects of disfluencies in naturalistic speech.

\paragraph{Effects of sentence length and parse on sentence-level syntactic information content.}
\newcommand{\length}[0]{\ensuremath{l}\xspace}
\newcommand{\depth}[0]{\ensuremath{d}\xspace}
% \alex{What do `pointwise estimates' mean?}
We inspect the Pearson correlation between the pointwise syntactic information content and two sequence-level features: length \length, as measured in number of words, and maximum tree depth \depth.
We conduct a separate correlation analysis for each dataset and each parse type (i.e., full parse and bracket-only).
% Visualizations of the relationships can be seen in \Cref{fig:followup_results}.

\begin{figure}[t]
    \centering
    \includegraphics[width=1\linewidth]{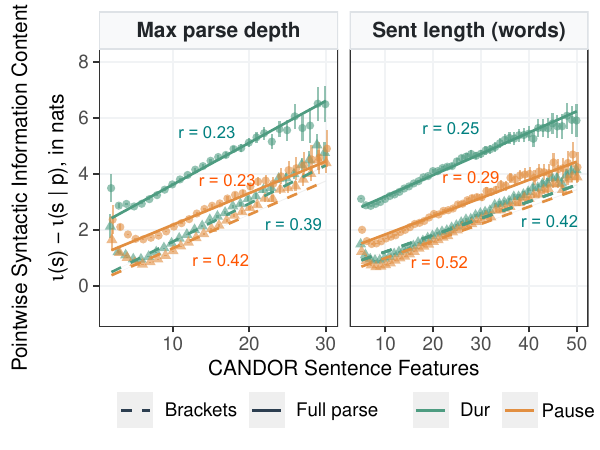}
    \caption{CANDOR sentence features versus pointwise syntactic information content. Each bin represents a single parse depth or sentence length value.
    % Error bars are 95\% confidence intervals for each bin.
    }
    \label{fig:followup_results}
\end{figure}

We find that \length and \depth are weakly inversely correlated with syntactic information content in pause and duration in planned speech ($|r|<0.05$ for all conditions, $p<0.001$ in all conditions but full parse information carried by pause).
We detail correlations in planned speech in Appendix \ref{sec:sentence-feature-analysis}.
In spontaneous speech, we observe a significant, moderate correlation for \length (pause: $r=0.52$; duration: $r=0.42$) and \depth (pause: $r=0.42$; duration: $r=0.39$), when considering phrase boundary information.
When considering full parses, the correlations are still significant, but slightly weaker for both \length (pause: $r=0.29$; duration: $r=0.25; $) and \depth (pause: $r=0.23$; duration: $r=0.23$).
For all reported $r$ values, $p<0.001$.
% Correlations on binned CANDOR data, shown in \Cref{fig:followup_results} are very high.
Overall, this analysis suggests that, when it comes to spontaneous speech, pause and duration contain greater phrase boundary information content in longer and syntactically deeper sentences.

\paragraph{Effects of sentence length on cross-genre trends.}
In \Cref{sec:results}, we find that prosody carries more syntactic information in spontaneous speech.
However, since length can directly influence the amount of sentence-level syntactic information, we perform a follow-up analysis controlling for sentence length.
We sample subsets of both speech datasets so that the sample includes the same number of sentences for each length in number of words ($5\leq l \leq 25$).
Even under exact length matching, spontaneous speech yields significantly higher syntactic information content for both duration ($3.59$ vs. $1.55$ nats; $p < .001$) and pause ($2.13$ vs. $1.94$ nats; $p < 10^{-17}$).

\paragraph{Effects of disfluencies in spontaneous speech.}
We conduct a post-hoc analysis to assess the prevalence of disfluencies in our data and their impact on our results.
While our data processing pipelines filter for sentence-initial disfluencies like restarts, we do not filter for sentence-medial disfluencies like repetitions and filled pauses, whose effects in our findings are unclear.
On one hand, prosody could be used to signal disfluency, making it easier for a listener to focus on the parts of the sentence that are strictly necessary for syntactic interpretation. 
On the other hand, disfluencies could add noise into the prosody stream, making relevant information harder to extract.
For this analysis, we define repetitions as identical adjacent tokens and filled pauses as the tokens \texttt{um}, \texttt{uh}, \texttt{er}, \texttt{ah}, and \texttt{hmm}.
We find that 19.8\% of our processed CANDOR sentences contain disfluencies.
While \mi is higher in disfluent subsets for all conditions, disfluent sentences are much longer ($\Bar{l}=17.9$ words) than fluent sentences ($\Bar{l}=11.99$ words).
A length-controlled analysis across fluency subsets shows that length is the primary driver of such difference, and disfluencies have a limited impact on our results.
We detail the analysis in Appendix \ref{sec:disfluency-analysis}.

\section{Discussion}
\label{sec:discussion}

\paragraph{Implications for linguistic theories.}

Our results support the mainstream linguistic theories of the syntax-prosody interface outlined in \Cref{sec:related-work}.
Starting with our two main theories of well-formedness, recall that Direct Reference Theory predicts a very high degree of informational overlap between prosody and syntax, whereas Indirect Reference predicts a moderate level.
Our results support indirect reference---pauses and duration \emph{do} carry syntactic information, but only up to about 10\% of the total.
Prosody-Driven Syntax predicts a higher degree of information in spontaneous speech over planned speech.
Our results are in line with this prediction: $\mi(\Prosody,\Syntax)$ values are higher in spontaneous speech, and $\uncert$ values are uniformly higher here, too, by up to $\approx6\%p$.
Finally, our results can be interpreted as strong support for prosodic bootstrapping.
We find that, even if a learner knows no words that are uttered in a sentence, they could still extract useful information about its syntax from prosody alone.

\paragraph{Prosody as an error-correcting code.}

What do our results suggest about the role of prosody in language comprehension, more broadly?
First and unsurprisingly, our results indicate that prosody largely signals phrase boundary information, and not phrase identities.
This corroborates recent research with similar findings \citep{degano2024speech}.
Of course, we did not study all prosodic features, and others, such as pitch, may carry more information about identity \citep[see e.g.,][]{wilcox-etal-2025-using}.
Second, prosody plays a largely redundant role for signaling syntax, when word identity is already known.
This suggests that, when it comes to syntax, prosody can be thought of as an error-correcting code \citep{mackay2003information}, i.e., information that can be used to reconstruct a message if the original signal is corrupted.
Given that different linguistic channels may be corrupted by different sources of noise, using a separate prosodic channel as a source of redundancy for important syntactic information may be optimal given resource constraints.
To this point, in \Cref{sec:follow-up-analysis}, we find that syntactic information content in disfluent spontaneous sentences is higher than that in fluent spontaneous sentences.
One possibility is that speakers adjust their prosodic usage in real-time to amplify its error-correcting role, for example to compensate for making disfluencies.
However, prosody as an error-correcting signal may not extend beyond syntax; recent work by \citet{yadavalli2025prosodytextconveycharacterizing} has shown that prosody contributes significant information above and beyond words for other tasks, such as emotion detection.

%However, our findings does not extend to ambiguity resolution.
%While prosody is also used for ambiguity resolution \citep{price1991use, Millotte2007phrasal}, this is a listener-side consideration.
%The framework in this work considers the speaker's side---that is, the relationship between prosody and syntax in a speaker’s utterance.
%The scope of our work does not include comprehension, so our results do not directly relate to ambiguity resolution.

\paragraph{Generalization beyond English.}
Our findings are from English data only; there is no guarantee that they generalize to other languages.
Given that pre-boundary lengthening and pause-based boundary marking are consistently observed cross-lingually \citep{Vaissiere1983language, fletcher2010prosody}, we predict that our high-level findings will hold when our methods are applied to multilingual datasets.
However, the redundancy between syntactic information carried by words and that by prosodic features may vary significantly cross-lingually, as phonological richness (e.g., tonality) or word order flexibility may affect how prosodic features are used to signal syntactic structure.

%Our findings that both duration and pause carry significant, measurable syntactic information in both planned and spontaneous speech, and that pause only carries significant syntactic information in planned speech supports all three camps of theories on the syntax--prosody interface, as each predicts some correspondence between syntactic and prosodic structure.

%However, we also find that pause in planned speech contains greater magnitude of syntactic information than in planned speech, which may be considered as evidence against Prosody-Driven Syntax.
%Under \prosdriv, which argues that the pressure to align syntax to prosody and vice versa is bi-directional, a speech style where only a uni-directional pressure is possible (planned read-out-loud speech) would result in a lower syntactic information content in prosody.
%\hyun{Todo: expand qualitative analysis with top highest pmi examples}

%Finally, we find that words already carry the same syntactic information carried by prosody.
%This supports IR theories that argue that prosody is also shaped by its own constraints that include word-specific properties (e.g. location of stress, number of syllables).
%Under IR, word identity contains information about word-specific phonological properties, thus rendering information from realized prosodic features redundant.

%While not a part of our research question, our experiments show that lexical identity reduces 44.2 nats of syntactic uncertainty in LibriTTS (71\% reduction), and 37.8 nats in CANDOR (86\% reduction).

\section{Conclusion}
In this work, we design and implement a pipeline to measure syntactic information content in two prosodic features: word duration and pause.
We confirm existing linguistic theories that prosodic features do carry measurable, statistically significant syntactic information content.
%In particular, word duration carries information about syntax in both planned and spontaneous speech, while pause only carries significant syntactic information in planned speech.
%We fail to find measurable information about syntax after controlling for what is already carried by words.
More broadly, this work presents novel, machine learning methods for measuring the interaction between prosody and syntax over large corpora.
We anticipate that our work will be useful for exploring a range of phenomena at the prosody-syntax interface, across a larger suite of datasets and languages.

\section*{Limitations}
\label{sec:limitations}
%\hyun{Should these points be integrated in the discussion section?}
%We organize our limitations into two large topics: improvements in experimental setup and improvements in prosody and text representation design.

\paragraph{Limitations with our approach.}
Our cross-entropy estimates of entropy, which we use to measure the syntactic information content via mutual information, are not unbiased estimates.
The better our model in predicting \Syntax, the lower the estimated entropy may be.
Thus, our estimates are an upper bound of the actual uncertainty in predicting \Syntax from \Words and \Prosody.
We consider the pre-trained text encoder to be an effective representation of \Words.
On the other hand, the prosody encoder which is trained from scratch and the fused representation of prosody and text encoder are likely less effective representations of \Prosody and \Words.
Thus, our estimates of $\ent(\Syntax \mid \Words)$ may be a much better estimate than our estimates of $\ent(\Syntax \mid \Words,\Prosody)$, leading to a deflated measurement of $\mi(\Syntax , \Prosody \mid \Words)$.
The negative \mi estimates in \Cref{tab:mutual-info} reflect this, as mutual informations are by definition always positive.\looseness=-1

Additionally, while we separately measure the syntactic information content in two features: duration and pause, it is possible that they contain redundant syntactic information.
Future work may benefit from combining the two features, or estimating the redundant information content between the two features.

Finally, we use gold transcripts for planned speech, versus ASR silver transcripts for spontaneous speech.
This may reduce the total amount of syntax predictable from text, and thus affect our mutual information estimates.
% This may underestimate the amount of syntactic information contained in prosodic features beyond text, and understate prosody's utility in noisy real-world or ASR settings.\looseness=-1

\paragraph{Model limitations.}
Despite the rising popularity of multimodality in language processing \cite{zhang-etal-2024-mm}, methods to reliably fuse text and real-valued input modes lag behind the likes of vision language models \citep{zhang2024vision, beyer2024paligemma} and speech and audio language models \citep{chu2023qwenaudioadvancinguniversalaudio, borsos2023audiolm}.
We believe our architecture can be further improved.
First, the internal architecture of our prosody encoder can be optimized further via hyperparameter search.
Our current encoders are fairly small and shallow, and it is possible that a deeper and larger encoder may be more effective.
In addition, earlier fusion of text and prosody input may also positively affect model performance, to allow interaction between the two modes at the encoder stage.
Second, although the output space of our experiments is extremely limited to 66 tokens representing POS and constituency labels and brackets, we currently utilize the entire pre-trained \citep{raffel2020t5} vocabulary.
This setup may result in probability leakage; the models may be assigning probabilities to illegal tokens during inference, adversely affecting our entropy estimates.
We investigate probability leakage in Appendix \ref{sec:leakage}.
Decoupling input and output tokenizers may help estimate entropy in syntax prediction more effectively.
Furthermore, models are known to be biased by tokenization, and to systematically assign less probability to strings with more tokens \citep{lesci-etal-2025-causal}; while we control for this in our spontaneous vs.\ read speech in \Cref{sec:follow-up-analysis}, this may affect the correlations in \Cref{fig:followup_results}.
Finally, the token embedding layer can also be decoupled between the encoder and the decoder.

\paragraph{Other experimental limitations.}
In our study, we investigate the effects of only two prosodic features---pause and duration.
However, there are many others, including pitch, loudness, tempo, and prominence, which is a composite prosodic feature.
%Both our selection and implementation of them can be improved.
%A future study can follow \citet{wolf-etal-2023-quantifying} to investigate additional prosodic features like prominence and pitch. 
%Loudness and tempo may also be features of interest.
% todo: add silver parse eval
In addition, our pipeline used silver parses, meaning that errors in the parse itself may be propagated and impact our results.
Finally, we only investigate English. The use of prosody is known to differ between languages, and running experiments on a wider set of languages is a necessary step to establish the generalizability of our results.

% \section*{Acknowledgment}
% This work was made possible by the Georgetown University High Power Computing's CLI Clusters.
% We thank the maintainers of the clusters Woonki Chung, Abhishek Purusthothama, and Dan DeGenaro.

\section*{Acknowledgements}
We thank Nathan Schneider, Amir Zeldes, members of the Georgetown University PICoL Lab, and anonymous reviewers for their help in shaping this work.
This work was made possible in part by support from Georgetown University HPC (managed by Woonki Chung), the CLI Leon Node (managed by Abhishek Purushothama and Dan DeGenaro), and the Digital Research Alliance of Canada / Alliance de recherche numérique du Canada.

\section*{Ethical Considerations}
Our work mainly concerns speech data, its transcriptions, and automatically generated annotations from NLP systems.
The datasets and NLP systems are publicly available; we do not report any potential issues regarding privacy or safety.
However, we acknowledge that our work is relevant to and contributes to research around large language models 
(LLMs) and other systems that are capable of generating and processing language data, possibly for malicious purposes.

Finally, we disclose the use of commercial LLMs as brainstorming, coding, and writing assistants during the design and implementation of the experiments and during the writing of this paper.

% Bibliography entries for the entire Anthology, followed by custom entries
%\bibliography{anthology,custom}
% Custom bibliography entries only
\bibliography{custom}

\newpage
\appendix
\begin{table*}[t]
\centering
\small
\setlength{\tabcolsep}{3pt}
\begin{tabular}{llcccccccc}
\toprule
 & & \multicolumn{4}{c}{\textbf{Uncontrolled}} & \multicolumn{4}{c}{\textbf{Length-controlled}} \\
\cmidrule(lr){3-6} \cmidrule(lr){7-10}
 & & \multicolumn{2}{c}{$(\Syntax, \Prosodypause)$} & \multicolumn{2}{c}{$(\Syntax, \Prosodyduration)$} & \multicolumn{2}{c}{$(\Syntax, \Prosodypause)$} & \multicolumn{2}{c}{$(\Syntax, \Prosodyduration)$} \\
\cmidrule(lr){3-4} \cmidrule(lr){5-6} \cmidrule(lr){7-8} \cmidrule(lr){9-10}
\textbf{Parse Type} & \textbf{Subset} & \multicolumn{1}{c}{\mi} & \multicolumn{1}{c}{\uncert} & \multicolumn{1}{c}{\mi} & \multicolumn{1}{c}{\uncert} & \multicolumn{1}{c}{\mi} & \multicolumn{1}{c}{\uncert} & \multicolumn{1}{c}{\mi} & \multicolumn{1}{c}{\uncert} \\ \midrule
\multirow{3}{*}{Brackets} & Fluent & 1.13 & 9.05\% & 1.36 & 10.9\% & 1.45 & 7.74\% & 1.68 & 8.96\% \\
 & Disfluent & 1.45 & 7.45\% & 1.62 & 8.33\% & 1.40 & 7.49\% & 1.58 & 8.42\% \\

 & Full & 1.20 & 8.64\% & 1.42 & 10.2\% & 1.43 & 7.62\% & 1.63 & 8.69\% \\
  \midrule
 \multirow{3}{*}{Full parse} & Fluent & 1.90 & 5.70\% & 3.31 & 9.92\% & 2.21 & 4.61\% & 3.69 & 7.70\% \\
 & Disfluent & 2.61 & 4.93\% & 3.91 & 7.39\% & 2.58 & 5.02\% & 3.88 & 7.55\% \\
 & Full & 2.05 & 5.52\% & 3.45 & 9.27\% & 2.40 & 4.82\% & 3.79 & 7.63\% \\ 
 \bottomrule
\end{tabular}
\caption{\mi in nats \uncert across fluent, disfluent, and full datasets in uncontrolled ($N = 298k$) and length-controlled ($N = 58k$ matched pairs, $\Bar{l}= 17.28$ words) CANDOR subsets. In each subset, \ent is computed separately, and the full ``subset'' rows are equivalent to those in \Cref{tab:mutual-info}.}
\label{tab:raw_sic_fluency_breakdown}
\end{table*}

\section{Probability Leakage}
\label{sec:leakage}
As discussed in \nameref{sec:limitations}, our model architecture \Cref{sec:model-arch} does not explicitly disallow illegal tokens while predicting syntactic parses.
When non-zero probabilities are assigned to illegal tokens (e.g. tokens that are not a POS tag, a phrase tag, or a bracket), our \ent estimates are inflated.
In this section, we analyze such probability leakage in our trained models.

For each trained model, we randomly select 1,000 sentences and measure at each prediction step the amount of probability assigned to illegal tokens.
The aggregate probability is then divided by the total number of prediction steps to obtain the average probability mass assigned to illegal tokens at a given prediction step.

We find that a significant probability mass is assigned to illegal tokens to full parse autoregressive models that estimate $\ent(\Syntax)$: 15.7\% in the planned speech (LibriTTS) model, and 11.7\% in the spontaneous speech (CANDOR) model.
Assuming a uniform distribution across $32k$ illegal tokens, the leakage contributes to $m_{leak}\times\ln{m_{leak}/N} = 1.92$ and $1.46$ nats of inflated \ent estimation, respectively, compared to an ideal model that assigns zero probability to illegal tokens.

The remaining models' leakage ranges from between 0.07\% (from  $\ent(\Syntax\mid\Prosodypause)$ estimation on CANDOR), accounting for around $0.01$ nats of inflation to 1.3\% (from $\ent(\Syntax\mid\Prosodyduration)$ estimation on LibriTTS), accounting for around $0.18$ nats of inflation. 
Models conditioned on brackets only leak no more than 0.001\% probability mass, producing negligible inflation.

\section{Modeling Hyperparameters}
\label{sec:hyperparams}
We employ k-fold cross-validation where $k=10$ for LibriTTS and $k=5$ for CANDOR, resulting in 90:10 and 80:20 train-test splits for each fold, respectively.
During training on each fold, we use the following hyperparameters: learning rate 3e-4 with a cosine scheduler, max length 256, batch size between 24 and 96, depending on GPU VRAM and modality, and early stopping patience 3.
Each LibriTTS experiment requires around 4 GPU-days on a single Tesla L4 GPU, while each CANDOR experiment requires around 2 GPU-days on a single NVIDIA H100 GPU.
Our models are trained to predit \Syntax by teacher-forcing.

\section{Details on Post-Hoc Analyses}

In \Cref{sec:follow-up-analysis}, we perform analyses into how sentence features (sentence length, parse depth) and disfluency affects our findings.
We provide additional detail on the two analyses in this section.

\subsection{Sentence Features and Sentence-level Syntactic Information Content in Planned Speech}
\label{sec:sentence-feature-analysis}

\begin{figure}[t]
    \centering
    \includegraphics[width=1\linewidth]{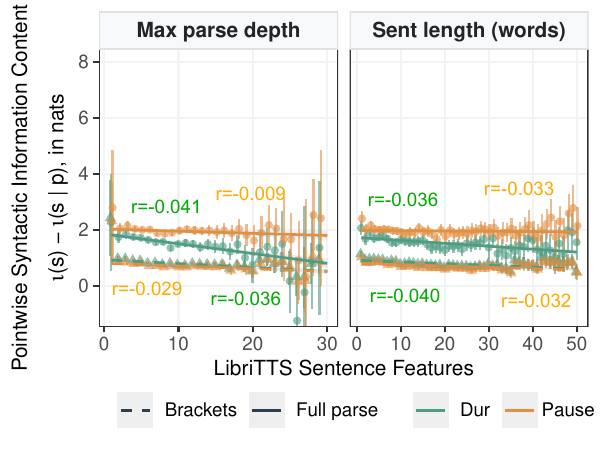}
    \caption{LibriTTS sentence features (sentence length and maximum tree depth) versus pointwise syntactic information content. Each bin represents a single depth or length value.
    % Error bars are 95\% confidence intervals for each bin.
    }
    \label{fig:sentence-features-libritts}
\end{figure}

We find that \length and \depth are weakly inversely correlated with sequence-level syntactic information content in pause and duration in planned speech.
The correlation is significant in all conditions when considering boundary information ($p<0.001$), with $r=-0.032$ for \length and pause, $r=-0.029$ for \depth and pause; $r=-0.040$ for \length and duration, $r=-0.036$ for \depth and duration.
The correlation is significant only in full parse information carried by duration with $r=-0.036, p < 0.001$ for \length and duration and $r=-0.041, p < 0.001$ for \depth and duration, while full parse information carried by pause is inversely, weakly, and insignificantly correlated: $r=-0.003$, $p=0.49$ for \length and pause, $r=-0.009, p=0.054$ for \depth and pause. 
We plot sentence features and sentence-level syntactic information content that duration and pause carry in planned speech in \Cref{fig:sentence-features-libritts}.

\subsection{Effects of Disfluency on $\mi(\Syntax,\Prosody)$}
\label{sec:disfluency-analysis}

At first glance, the fluent and disfluent subsets vary significantly in \mi and \uncert, as seen in \Cref{tab:raw_sic_fluency_breakdown}.
For example, in the disfluent subset, duration has $3.91$ nats of \mi with syntax, compared to $3.31$ in the fluent subset. And pauses $2.61$ nats of \mi, vs. only $1.90$ in the fluent subset.

However, given the large difference in mean sentence length $\tilde{l}$ across the fluent (11.99 words) and disfluent (17.90 words) subsets, the length could be the primary driver in the cross-fluency variation.

A length-controlled analysis, similar to our cross-genre analysis in \Cref{sec:follow-up-analysis}, shows that \length accounts for the majority of this difference, reducing the cross-subset gap in both \mi and \uncert.
After controlling for \length, the difference in \mi between the subsets and the full dataset drops from up to $40\%$ of the total value to within $7.5\%$ of the total value.

To statistically confirm that \length is in fact the main driver for the uncontrolled cross-subset difference, we perform a regression analysis by fitting ordinary least squares models.
We compare a baseline model predicting $\Delta\surp = \surp(\syntaxunit) - \surp(\syntaxunit,\prosodyunit)$ using only \length, against an augmented model containing both \length and binary fluency information (fluent or disfluent).
Across all conditions, adding fluency information to a regression model provides little added explanatory power, with $\Delta R^2 < 0.01$.
That is, length is the main driver for the variance of sentence-level syntactic information content between fluent and disfluent subsets.

\end{document}